\documentclass[10pt,conference]{IEEEtran}

\AtBeginDocument{%
  \providecommand\BibTeX{{%
    \normalfont B\kern-0.5em{\scshape i\kern-0.25em b}\kern-0.8em\TeX}}}

\usepackage[framemethod=tikz]{mdframed}
\usepackage{lipsum}
\usetikzlibrary{shadows}
\usepackage{mathtools}
 \usepackage{paralist}
 \definecolor{light-gray}{gray}{0.80}
\usepackage{graphics}
\usepackage[shortlabels]{enumitem}
\usepackage{dblfloatfix}
\usepackage[utf8]{inputenc}
\usepackage{multicol}
\usepackage{graphicx}
\usepackage{wrapfig}
\usepackage{colortbl}
\usepackage[ruled,vlined]{algorithm2e}
\usepackage[final]{listings}
\usepackage[framemethod=tikz]{mdframed}

\usepackage{lipsum}
\usetikzlibrary{shadows}
\usepackage{mathtools}
 \usepackage{paralist}
 \definecolor{light-gray}{gray}{0.80}
\usepackage{graphics}
\usepackage[shortlabels]{enumitem}
\usepackage{dblfloatfix}
\newmdenv[
  tikzsetting= {fill=light-gray},
  linewidth=1pt,
  roundcorner=0pt, 
  shadow=false
]{myshadowbox}
\usepackage{colortbl} 
\usepackage{blindtext, graphicx}
\usepackage{textcomp}
\usepackage[hidelinks]{hyperref}
\usepackage{amsmath}
\usepackage{amsfonts}
\usepackage{caption}
\usepackage{color}
\usepackage[final]{listings}
\usepackage{balance}
\usepackage{picture}
\usepackage{relsize}
\usepackage{multicol}
\usepackage{soul}
\usepackage{enumitem}
\setitemize{noitemsep,topsep=0pt,parsep=0pt,partopsep=0pt,leftmargin=*}
\setenumerate{noitemsep,topsep=0pt,parsep=0pt,partopsep=0pt,leftmargin=*}
\usepackage{tikz}

\usepackage{makecell}
\usepackage{bigstrut}
\usepackage{amsmath,amssymb,amsfonts}
\usepackage{xspace}
\usepackage{graphicx}
\usepackage{booktabs}
\usepackage{comment}
\usepackage{threeparttable}
\usepackage{subcaption}

\begin{document}

\newcommand{\etal}{\textit{et al.}\xspace}

\newcommand{\pkfocus}{\texttt{PK-FoCus}\xspace}
\newcommand{\ourmethod}{\texttt{PK-NCLI}\xspace}
\newcommand{\real}{\mathbb{R}\xspace}
\newcommand{\aggr}{\texttt{aggr}\xspace}
\newcommand{\context}{\texttt{CTXT}\xspace}
\newcommand{\candidate}{\texttt{CAND}\xspace}
\newcommand{\softmax}{\texttt{softmax}\xspace}
\newcommand{\simi}{\texttt{sim}\xspace}
\newcommand{\identity}[1]{\mathbb{I}({#1})\xspace}
\title{Context Retrieval via Normalized Contextual Latent Interaction for Conversational Agent}




\author{
\IEEEauthorblockN{
Junfeng Liu$^{1,2}$,
Zhuocheng Mei$^{1}$, 
Kewen Peng$^{1}$, and
Ranga Raju Vatsavai$^{1,3}$}
\IEEEauthorblockA{
$^{1}$
\textit{Department of Computer Science, North Carolina State University}, 
Raleigh, NC, USA \\
$^{2}$
\textit{Lirio AI Research, Lirio LLC}, 
Knoxville, TN, USA\\
$^{3}$
\textit{Behavior Reinforcement Learning Lab, Lirio LLC}, 
Knoxville, TN, USA\\
\{jliu85, zmei5, kpeng, rrvatsav\}@ncsu.edu
}
}

\maketitle

\begin{abstract}

Conversational agents leveraging AI, particularly deep learning, are emerging in both academic research and real-world applications. However, these applications still face challenges, including disrespecting knowledge and facts, not personalizing to user preferences, and enormous demand for computational resources during training and inference. Recent research efforts have been focused on addressing these challenges from various aspects, including supplementing various types of auxiliary information to the conversational agents. However, existing methods are still not able to effectively and efficiently exploit relevant information from these auxiliary supplements to further unleash the power of the conversational agents and the language models they use. In this paper, we present a novel method, \ourmethod, that is able to accurately and efficiently identify relevant auxiliary information to improve the quality of conversational responses by learning the relevance among persona, chat history, and knowledge background through low-level normalized contextual latent interaction. Our experimental results indicate that \ourmethod outperforms the state-of-the-art method, \pkfocus, by 47.80\%/30.61\%/24.14\% in terms of perplexity, knowledge grounding, and training efficiency, respectively, and maintained the same level of persona grounding performance. We also provide a detailed analysis of how different factors, including language model choices and trade-offs on training weights, would affect the performance of \ourmethod.

\end{abstract}
\section{Introduction}
Recent advances in machine learning and deep learning have enabled a tremendous amount of applications in natural language processing, particularly in conversational AI. Many dialogue agents are driven by conversational data. While they can deliver a reasonable answer, many existing methods fail to address important auxiliary information that is relevant to the conversation, leading to an answer irrelevant to the topic. For instance, in a question-answering system, when an agent is processing a query from a human without external knowledge related to the topic, it is nearly impossible to construct an ideal answer based on facts. Hence, to provide accurate and relevant answers it is critical that an AI agent should leverage external knowledge and respect such knowledge in the generated responses.

Another emerging need in conversational AI applications is to seek personalized answers so that these answers could help improve user experience and engagement during the conversation. Many existing research efforts have been focused on leveraging the speakers' persona information to understand the user's intent and tailor the responses to the specific user. 

In this paper, we focus on incorporating both external knowledge and user persona to improve the quality of answers generated by conversational models.  While knowledge provides background information about discussion topics and other ground truth information, it is not guaranteed to generate the response that the user seeks. Without capturing the user's intention or preference, it is a challenge for a model to satisfy the user's need. By learning the user's persona, a model can decide which knowledge a user is looking for. Therefore, learning the user's persona is critical for a human-like conversation. 
We propose a novel framework, \ourmethod, with knowledge and persona grounding based on normalized contextual latent interaction that helps identify relevant knowledge and persona entries to improve the quality of responses generated by two benchmark language models. 
Our experiments demonstrate that \ourmethod is able to outperform the state-of-the-art persona/knowledge-grounding method, \pkfocus~\cite{jang2022call}, by 47.80\% in terms of language quality, 30.61\% in terms of knowledge grounding, and maintain the same persona grounding performance. 
In addition, our method \ourmethod improved training efficiency over the baseline by 24.14\%.


\section{Related Work}
With the increasing computational capability of deep neural networks (DNNs),  many researchers endeavor to develop AI agents for automated dialog generation for various applications, such as question answering  and machine translation.
However, one of the major challenges is to generate informative yet diverse conversations that are suitable in a specific context. One approach is to utilize personalized information (e.g., demographics, hobbies, preferences) to train chatbots that can generate customized conversations. 
Li~\etal~\cite{li2016persona} proposed a persona-based dialog generation model, which is capable of generating responses that are consistent with a particular persona. 
Similarly, Humeau~\etal~\cite{humeau2019poly} proposed an approach in which the neural network is trained on a personality-capturing loss function such that the generated conversation is more likely to align with certain given personality traits. Other recent works also work toward similar directions in designing persona-inclusive dialog models~\cite{zhang2018personalizing,zheng2020pre}.
Recently, Jang~\etal~\cite{jang2022call} created a conversation dataset that is supported by both speaker personas and external knowledge base and proposed a novel persona- and knowledge-grounding method to generate novel responses. 

\subsection{Neural-Based Model}
Traditional conversation AI agents require specifically engineered tools, such as a carefully crafted knowledge graph, external API calls, etc. They also rely heavily on human expertise for evaluation. These constraints largely limit the applicability and scalability of the agents to be widely deployed. Neural networks and data-driven approaches are able to overcome these limits by extracting signals directly from data through end-to-end training without human experts, and their powerful capabilities have been expanded to wider applications.
These neural-based models can be further categorized as retrieval-based approaches and generation-based approaches. Retrieval-based models select a response from the pool of candidates by learning similarities between the input query and candidates. 
Bi-encoders~\cite{lee2019contextualized,reimers2019sentence} and cross-encoders~\cite{devlin2018bert,liu2019roberta} are two popular approaches
to learn query-candidate similarities. Bi-encoders typically encode the input query and the response candidates into a lower-dimension space and calculate the similarity by certain distance metrics, while cross-encoders typically pre-fuse the query and candidates together (e.g., text concatenation) and learn a joint embedding before generating a score based on the joint embedding. Humeau~\etal~\cite{humeau2019poly} proposed a poly-encoder, which combines the advantages of both bi-encoder and cross-encoder. 
Many existing information retrieval solutions can also be easily adapted to ranking-based conversation models when the candidate responses are viewed as candidate documents. 
The state-of-the-art retrieval method, ColBERT~\cite{khattab2020colbert}, leverages contextual latent interactions to better capture the low-level word similarities between the query and documents. They improve the computational performance over the BERT~\cite{devlin2018bert} model by pre-computing the document embeddings. 
%
Beyond the cost challenge, retrieval-based models struggle to provide creative and novel responses beyond the prescribed candidates from the dataset. Compared to retrieval-based models, generation-based models are able to provide more novel and creative responses by generating sequences of tokens as the response instead of selecting from an existing dataset. 
%
%
Recurrent networks~\cite{li2016persona} and auto-regressive  networks~\cite{hoogeboom2021autoregressive, oord2016wavenet} are two popular frameworks for generation tasks. The generation models are typically learned by optimizing the probability of the next word over a set of words from a vocabulary. 

Substantial research efforts have been made towards exploring auxiliary information that can be used to improve the performance of a conversational agent, including speaker personas~\cite{li2016persona, zhang2018personalizing}, 
visual environments~\cite{mostafazadeh2017image},
external knowledge~\cite{ghazvininejad2018knowledge, dinan2018wizard}, etc.

\subsection{Knowledge-based Model}
Without sufficient knowledge background, a generated response might still not be desired even though it is fluent, as the response might not respect true knowledge. Ghazvininejad~\etal~\cite{ghazvininejad2018knowledge} introduced a neural network approach that matches dialogue history with world facts, and fed it into a neural architecture to generate responses. Such approach aims to map the existing knowledge to the conversation topic for a more accurate response. Dinan~\etal~\cite{dinan2018wizard} developed a conversational agent powered by knowledge by leveraging the Generative Transformer Memory Network. The knowledge candidates are first selected by an information retrieval model, and then are encoded along with dialog context by a transformer in a two-stage end-to-end network. The decoder generates the responses by attending over both the knowledge and dialogue. 
Empirically, knowledge-based agents are not only able to outperform other approaches without knowledge background, but they are generalizable to topics that are not present during the training stage.

\subsection{Persona-based Model}
A particular line of research focuses on leveraging speaker personas to improve the personalization level of the generated responses in various applications. A generic conversational model trained without persona may be able to produce legitimate responses, but such responses might not be tailored or align with the speaker's intention~\cite{liu2022persona}. Hence, speaker personas are critical to providing the most appropriate responses, especially in applications where personalization is highly desired.
Many existing methods~\cite{humeau2019poly} treat persona as a part of the input query through concatenation. However, these approaches are limited to only text-based personas. Moreover, persona and query are two different sources of information, and processing them heterogeneously might bring more complexity to the learning problem and result in lower performance.
To tackle this challenge, Zhang \etal~\cite{zhang2018personalizing} introduced a persona-based conversation dataset and proposed the Profile Memory Network (\textit{PMN}) that encodes the persona profiles into conversation context for both generation-based and retrieval-based response generation tasks. \textit{PMN} deploys Seq2Seq model for generating responses. The encoder takes in the encoded dialogue history with Long Short-Term Memory (LSTM), and the decoder attends over the encoded profile entries. 
Liu~\etal~\cite{liu2023pcpe} developed a persona-based ranking model that encodes persona and query in multiple encoding stream and learns a context-aware embedding via a post-fusion approach. 
\section{Methodology}
\label{sec:method}

\subsection{Problem Definition}
\label{sec:method:problem}
We consider a conversation generation problem that leverages a language model based on a user's persona and ground knowledge. Formally, given a conversation history $U$ (of which the last utterance is a question asked by a user), together with the user's persona $P$ with $N_p$ entries $ P = \{P_1, \dots, P_{N_p}\}$ and a ground knowledge $K$ with $N_k$ paragraphs $K = [K_1, \dots, K_{N_k}]$, we want to train a model $f$ that is able to generate an answer $a$ such that $f: (U, P, K) \rightarrow a$.
In this setup, the question $q$, answer $a$, persona entries $P_1, \dots, P_{N_p}$ and ground knowledge paragraphs $K_1, \dots, K_{N_k}$ are all texts. 

\begin{figure*}[t!]
    \centering
    \includegraphics[width=0.65\linewidth]{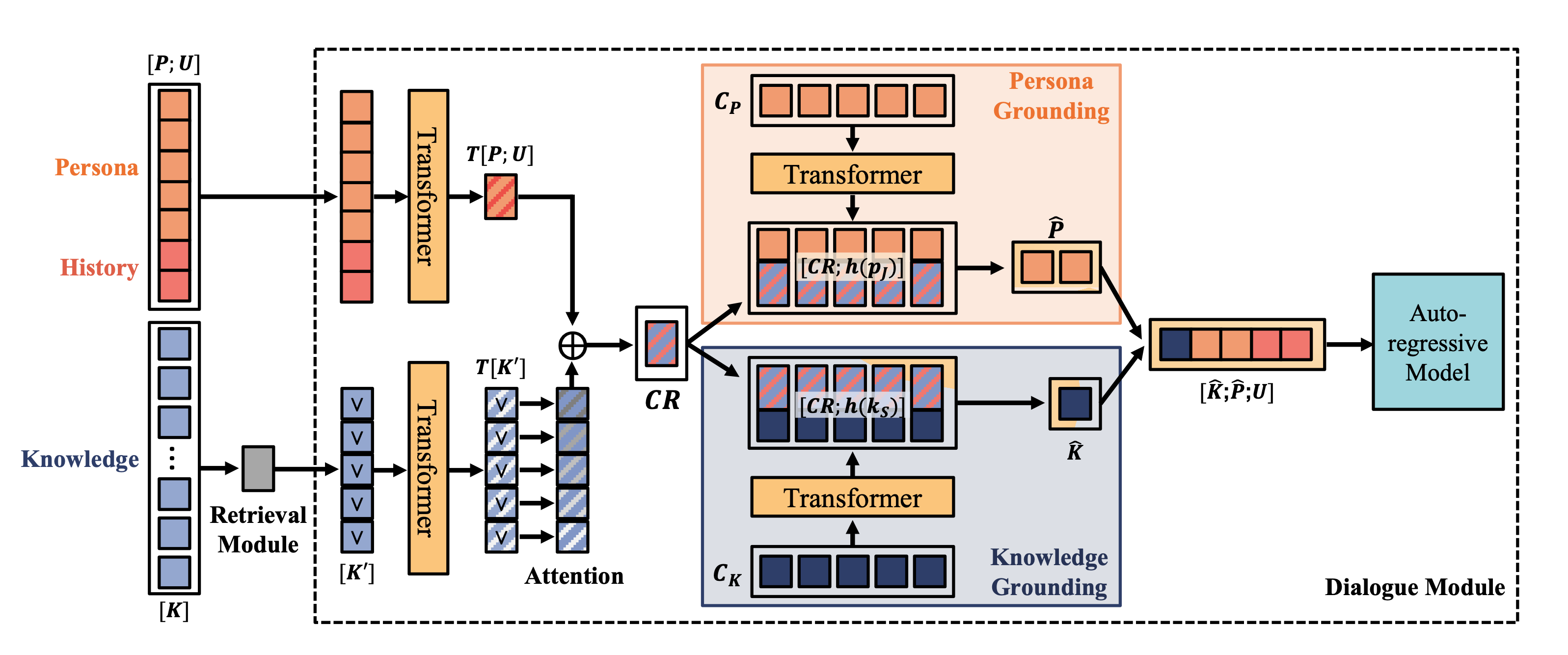}
    \caption{Network Architecture of \pkfocus~\cite{jang2022call}}
    \label{fig:network-focus}
    \vspace{-10pt}
\end{figure*}

\subsection{Baseline Method}
\label{sec:method:baseline}
We consider \pkfocus~\cite{jang2022call} by Jang~\etal as our baseline, where they fine-tuned existing language models (such as GPT-2~\cite{radford2019gpt2} and BART~\cite{lewis2019bart}) with persona grounding (PG) and knowledge grounding (KG) blocks that identifies relevant persona and knowledge to the conversation. 
Specifically, \pkfocus used the above language models to encode persona and conversation history $[P; U]$ and knowledge $K$, respectively. 
A context-relevant representation $CR$ is generated based on the embeddings of $[P; U]$ and $K$. 
$CR$ is further used in the PG block to sample a subset of relevant persona entries $\hat{P}$
and in the KG block to sample one relevant knowledge $\hat{K}$. 
Then $\hat{P}$, $\hat{K}$ and the history $U$ are concatenated as the input to the BART model for language encoding and decoding. The decoded output is the answer generated by the model in response to the user's question. 
Figure~\ref{fig:network-focus} shows the network architecture of the baseline Persona and Knowledge Grounding. 

While \pkfocus presented a new dataset that combines both knowledge and persona in the conversation and a novel method that identifies relevant persona/knowledge entries to supplement the conversation context, it also suffer from several aspects. 
Firstly, \pkfocus uses a concatenation-based fusion on various inputs during the persona/knowledge grounding stage, which has been proven to be suboptimal in many existing research~\cite{liu2023pcpe}. 
Secondly, language embedding is expensive, especially with large language models (LLMs) like GPT2. \pkfocus uses the language model encoders four times in the grounding process, while the embeddings could not be reused (i.e., $T[P; U]$ could not be reused as persona embedding $T[P]$ in persona grounding stage, and similarly for knowledge grounding). This procedure is very computationally expensive, and could be critical to real applications when a fast response is desired.

To address the above issues of \pkfocus, we propose our method, \ourmethod, which can leverage low-level word-interaction to better learn the relevance between the persona/knowledge entries and the conversation context, while providing a framework that is more computationally efficient. 
In addition, we have corrected several implementation errors in the source code provided by the baseline \pkfocus, including tokenizer and evaluation metrics.

\subsection{Our Method}
\label{sec:method:our}

\begin{figure*}[t!]
    \centering
    \includegraphics[width=0.75\linewidth]{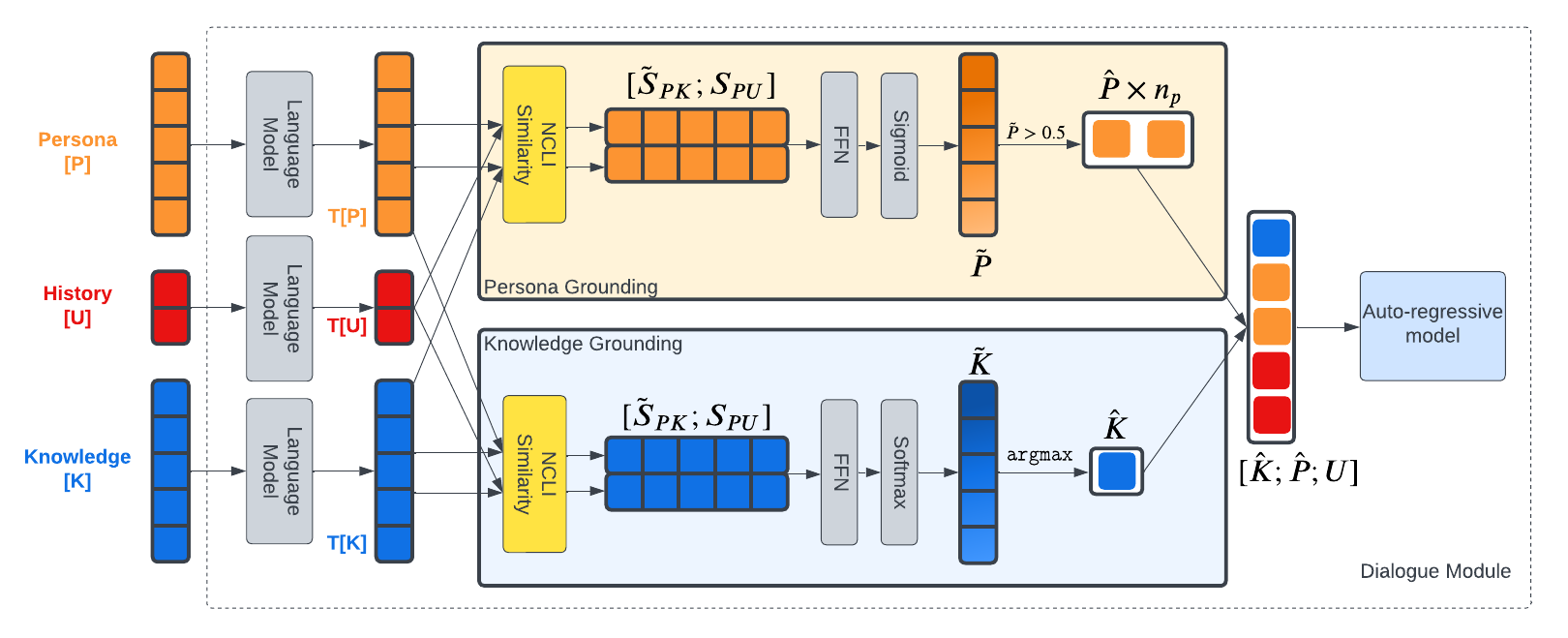}
    \caption{Overview of \ourmethod Model Architecture}
    \label{fig:our-approach}
\end{figure*}

Our method, Persona and Knowledge Chat with Normalized Contextual Latent Interaction (denoted as \ourmethod), is demonstrated in Figure~\ref{fig:our-approach}. 
From higher level, \ourmethod is given three sources of inputs ($P, K, U$), and it attempts to identify relevant persona and knowledge entries through improved PG and KG processes ($\hat{P} \in P$ and $\hat{K} \in K$) and amend them to the original utterance ($U$) so that a simultaneously trained auto-regressive language model can generate a response. 
In the grounding processes, \ourmethod only encodes the three input sources once and reuses for both PG and KG, then it leverages low-level contextual latent interaction among the three sources to better explore the relevance of the corresponding inputs.



\subsubsection{Inputs Embedding}
\label{sec:method:our:input}
The first step of \ourmethod is to embed the input utterance, candidate persona and knowledge entries. In this step, \ourmethod could use any existing language embedding models that convert from the vocabulary space to language embedding vectors. 
The language model ($LM$) embeds the utterance, persona entries, and candidate knowledge entries, respectively, and produces word-level embedding vectors as 
\begin{equation}
\label{eqn:input-emb}
\begin{aligned}
    T[U] & = LM(\text{U}),  \\
    T[P] & = LM(\text{P}),  \\
    T[K] & = LM(\text{K}),  \\
\end{aligned}
\end{equation}
where 
$T[U] \in \real^{b \times 1 \times s \times d}$, 
$T[P] \in \real^{b \times N_p \times s \times d}$,
$T[K] \in \real^{b \times N_k \times s \times d}$, 
$b$ is the batch size, $s$ is the sequence length (number of words in the text) of the corresponding input, $d$ is the embedding dimension, and $N_p$/$N_k$ are the numbers of candidate persona/knowledge entries of one conversation. 


In our experiments, we use GPT-2~\cite{radford2019gpt2} and BART~\cite{lewis2019bart} as the language embedding models, following the work of Jang~\etal~\cite{jang2022call}, as these two models are the leading transformer-based and RNN-based language models, respectively.

\subsubsection{Normalized Contextual Latent Interaction}
\label{sec:method:our:latent-sim}
In a conversation, not all supplemental information is relevant to the context and thus might not be
useful for response generation.  
In order to effectively identify the persona/knowledge entries that are relevant to the conversation,
we propose a Normalized Context Latent Interaction (NCLI), inspired by ColBERT~\cite{khattab2020colbert},
that leverages low-level word similarities so that the model can make informed choices of the relevant persona/knowledge.

\begin{figure}[t!]
    \centering
    \includegraphics[width=0.9\linewidth]{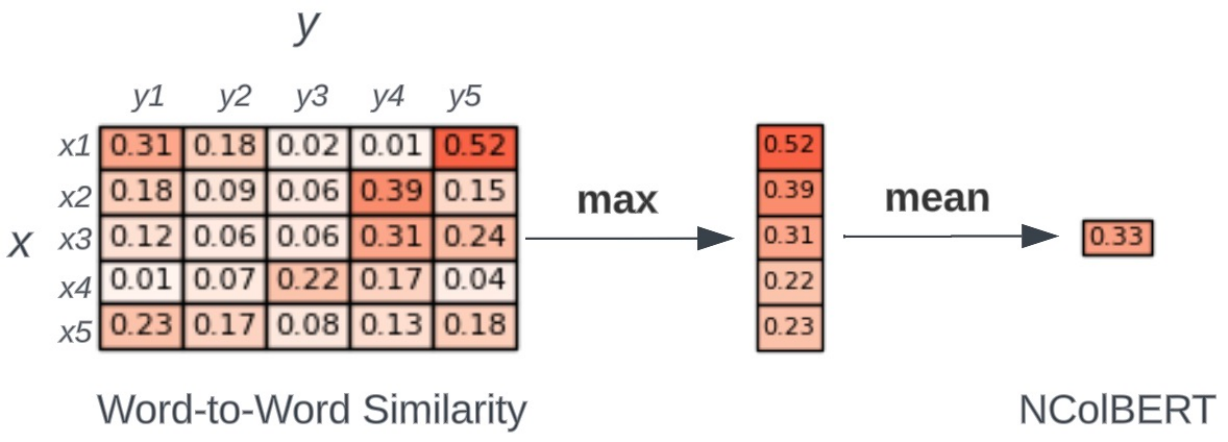}
    \caption{Demonstration of $\texttt{NColBERT}$}
    \label{fig:ncolbert}
\end{figure}

We first introduce the normalized ColBERT ($\texttt{NColBERT}$) score between two sentences $x$ and $y$ (shown in Figure~\ref{fig:ncolbert}). 
Here we consider a slightly different ColBERT similarity calculation from~\cite{khattab2020colbert}, which is normalized by the length of $x$, i.e., 
\begin{equation}
\label{eqn:ncolbert}
    \texttt{NColBERT}(x, y) = \frac{1}{|x|} \sum_{x_i \in x} \max_{y_j \in y} E_{x_i} \cdot E_{y_j}^T,
\end{equation}
where $x_i$ is the $i$-th word in sentence $x$,
$y_j$ is the $i$-th word in sentence $y$,
$E_{x_i}$/$E_{y_j}$ are the embeddings of $x_i$/$y_j$,
and $|x|$ is length of $x$. 
The $\texttt{NColBERT}$ similarity is different 
from the original ColBERT calculation by the normalization term $\frac{1}{|x|}$. In practice, we identified an issue with the original ColBERT that longer sentences $x$ are likely to dominate the similarity distributions among other inputs because of the sum operation, regardless of the low-level matching by word similarity. This issue was not addressed in~\cite{khattab2020colbert} as the original ColBERT was designed for document retrieval tasks, where there is only one query (i.e., $|x|=1$). In our application where we want to measure the similarity between multiple $x$'s and multiple $y$'s (i.e., multiple persona/knowledge entries), such normalization becomes critical to eliminate the effect from varying lengths of the sentences and only focus on low-level information match.

Next, we define $\texttt{NCLI}(X, Y)$ as the pairwise similarity function for entries from two different sources $X$ and $Y$ using $\texttt{NColBERT}$.
In our application, we consider four tuples of the inputs $(X, Y)$: 1. persona and utterance, 2. persona and knowledge, 3. knowledge and utterance, and 4. knowledge and persona. And their pairwise $\texttt{NCLI}$ are
\begin{equation}
\label{eqn:NCLI}
\begin{aligned}
    S_{PU} & = \texttt{NCLI}(T[P], T[U]) \in \real^{N_p \times 1}, \\
    S_{PK} & = \texttt{NCLI}(T[P], T[K]) \in \real^{N_p \times N_k}, \\
    S_{KU} & = \texttt{NCLI}(T[K], T[U]) \in \real^{N_k \times 1}, \\
    S_{KP} & = \texttt{NCLI}(T[K], T[P]) \in \real^{N_k \times N_p}, \\
\end{aligned}
\end{equation}
where $S_{XY}$ is the pairwise $\texttt{NCLI}$ similarity matrix between the input $X$'s entries and input $Y$'s entries.

Since $\texttt{NColBERT}$ calculates low-level pairwise similarities on word level among the inputs
and thus is computationally expensive. Therefore, the input embeddings from Equation~\ref{eqn:input-emb}
are first reduced to a lower dimension $d_0$ ($d_0 < d$) before feeding to the ColBERT similarity calculation in Equation~\ref{eqn:ncolbert} to improve training and testing efficiency. In our experiments, we use $d_0 = d/4$. 

Note that in $\texttt{NColBERT}$ is an asymmetric similarity measure ($max$ along $y$'s). Consequently, $\texttt{NCLI}$ is also asymmetric. That is, $\texttt{NColBERT}(x, y) \ne \texttt{NColBERT}(y, x)$ and $\texttt{NCLI}(X, Y) \ne \texttt{NCLI}(Y, X)$.

\subsubsection{Persona Grounding}
\label{sec:method:our:pg}

In the previous step, NCLI calculates the low-level similarity between two inputs.
Here in the persona grounding (PG) layer, we explain how we use these low-level similarities
to identify which persona entries are relevant to the conversation. 
Specifically, we calculate the average similarity between a persona entry and the knowledge entries 
$\tilde{S}_{PK} = \frac{1}{N_k} \sum_{i=1}^{N_k} S_{PK_i}$. 
Then the persona's similarity score with the knowledge $\tilde{S}_{PK}$ and with the utterance $S_{PU}$ are fused together by a feed-forward layer (FFN) and followed by a sigmoid activation as 
\begin{equation}
\tilde{P} = \sigma(w_1 \tilde{S}_{PK} + w_2 S_{PU} + b).
\end{equation} 
where $w_1$/$w_2$ are trainable weights, $b$ is a trainable bias, and
$\tilde{P} \in \real^{b \times N_p}$ represents probability of whether a persona entry $P_i$ is relevant. The selected persona entry $\hat{P}$ is determined by the probability greater than 0.5, i.e., 
\begin{equation}
    \hat{P} = P[\tilde{P} > 0.5].
\end{equation}

\subsubsection{Knowledge Grounding}
\label{sec:method:our:kg}

A similar knowledge grounding (KG) layer is adopted to identify relevant knowledge entries for the conversation.
The average knowledge-persona similarity $\tilde{S}_{KP}$ is calculated as 
$\tilde{S}_{KP} = \frac{1}{N_p} \sum_{i=1}^{N_p} S_{KP_i}$. The knowledge entry probabilities $\tilde{K}$ 
and the relevant knowledge entry $\hat{K}$ are obtained by
\[
\begin{aligned}
    \tilde{K} & = \softmax(w_1 \tilde{S}_{KP} + w_2 S_{KU} + b), \\
    \hat{K} & = K[argmax(\tilde{K})].
\end{aligned}
\]



Note that the knowledge grounding is slightly different from the persona grounding in terms of the number of selections. We follow this design from \cite{jang2022call} and assume that in one utterance, there could be multiple relevant persona entries, but the responses should be based on exactly one knowledge entry. 
Correspondingly, softmax activation is used in the KG layer instead of sigmoid activation.

\subsubsection{Response Generation}
\label{sec:method:our:generation}

To generate the response to the original given utterance, we utilize the language model ($LM$) from Section~\ref{sec:method:our:input}, following many existing generation-based conversation models~\cite{jang2022call, radford2019gpt2}. 
The selected knowledge entry $\hat{K}$ and persona entries $\hat{P}$ are concatenated with the utterance $U$ as the input to $LM$, i.e., 
\begin{equation}
    O = LM([\hat{K}; \hat{P}; U]).
\end{equation}
The output embeddings $O$ are then projected to a probability distribution $p_r$ over the vocabulary space. 

For training, the model is trained on the loss function below:
\begin{equation}
    L = \alpha L_{KG} + \beta L_{PG} + \gamma L_{LM}
\end{equation}
where $L_{KG}$ is the knowledge grounding loss based on the cross-entropy over $\tilde{K}$, 
$L_{PG}$ is the persona grounding loss based on the cross-entropy over $\tilde{P}$, 
$L_{LM}$ is the language model loss, which is the cross-entropy loss over the ground 
truth response and the probability distribution $p_r$, and $\alpha$/$\beta$/$\gamma$ are the weights
for the $L_{KG}$/$L_{PG}$/$L_{LM}$ terms, respectively.

For inference, $LM$ follows the auto-regressive~\cite{oord2016wavenet, hoogeboom2021autoregressive} framework to generate a sequence of tokens as the response. It first samples one token from $p$, and appends to the input $[\hat{K}; \hat{P}; U]$ for the next token generation until the \texttt{<EOS>} (end-of-sentence) token is sampled or maximum length is reached. 

\section{Experiments}
\label{sec:experiments}

\begin{table}[t]
\centering
\caption{Statistics of the FoCus dataset}
\begin{tabular}{lrr} \toprule
                              & \multicolumn{1}{c}{Train} & \multicolumn{1}{c}{Valid} \\ \midrule
\# Dialogs                    & 12484                     & 1,000                     \\
\# Average Rounds             & 5.63                      & 5.64                      \\
Avg. Length of Human's Utt.   & 40.70                     & 40.21                     \\
Avg. Length of Machine's Utt. & 138.16                    & 138.60                    \\
\# Knowledge-only Anser       & 37,488                    & 3,007                     \\
\# Persona-Knowledge Answer   & 32,855                    & 2,630                     \\
\# Landmarks                  & 5,152                     & 923         \\ \bottomrule             
\end{tabular}
\label{tab:dataset}
\end{table}

\subsection{Dataset}
\label{sec:experiments:dataset}
We use the FoCus dataset provided by Jang~\etal~\cite{jang2022call}. 
The FoCus dataset was generated by crowd-sourced workers from Amazon Mechanical Turk. 
Each worker was requested to generate conversations that involved a user and a chatbot chatting in turns based on given persona entries of the user and ground truth knowledge of the conversation topic.  
Each user is assigned five persona entries, each of which is a sentence that describes the user's personality, such as hobbies. 
The ground truth knowledge of a conversation topic is the text extracted from a Wikipedia page about a landmark. 
The dataset consists of 13,484 conversations, among which 12,484 are for training, and 1,000 are for validation. In the training set, each conversation has, on average, 5.63 turns that involve 5,152 unique landmarks. Note that all answers are more or less based on knowledge, but not all answers are based on personas. Answers in 37,488 turns are based on only knowledge, and answers in 32,855 are based on persona and knowledge. 
More detailed statistics are available on the FoCus dataset page~\footnote{\url{https://github.com/pkchat-focus/FoCus}}.

\subsection{Experimental Setup}
\label{sec:experiments:setup}
To empirically assess the viability of our approach, we have conducted experiments centering around the following three research questions:

\begin{itemize}
    \item {\bf RQ1:} Can the \texttt{NCLI} effectively extract relevant information and identify relevant persona/knowledge entries relevant to the conversation context?
    \item {\bf RQ2:} Performance-wise, is it helpful to adjust weights on different loss functions and use hyperparameter optimization?
    \item {\bf RQ3:} Is \ourmethod more computationally efficient than the baseline \pkfocus?
\end{itemize}

For evaluation purposes, we consider the following evaluation metrics. The perplexity score (\textbf{PPL}) measures the statistical confidence of the model about the predicted texts. \textbf{ROUGE} and \textbf{BLUE} scores indicate semantic textual similarity between two sentences. The \textbf{F1} score is the binary classification metric derived from the confusion matrix. \textbf{PG}/\textbf{KG} accuracy measures the accuracy of selecting the correct candidates. Since there are zero or more persona entry labels for each conversation, we also report a multi-label version of PG accuracy (PG\_MTL), which requires all persona entry labels to be predicted correctly.

To answer {\bf RQ1}, we evaluate \texttt{NCLI} by the PG/KG accuracy. 
To answer {\bf RQ2}, we evaluate the quality of the recommended candidates by measuring perplexity scores (PPL), ROUGE scores, and BLUE scores. 
To answer {\bf RQ3}, we evaluate the training/inference time of the baseline versus our proposed approach on two language models (BART and GPT-2), respectively. We also provide an analysis of \ourmethod framework's efficiency on potential practical applications in Section~\ref{sec:results:rq3}. 

Our method is implemented following the framework provided by Jang~\etal~\cite{jang2022call}. The source code of our method is available on Github~\footnote{\url{https://github.com/jliu-v/pk-ncli}}. 
In addition to the default setup ($\alpha/\beta/\gamma=1/1/10$) suggested by~\cite{jang2022call}, we conducted hyper-parameter search on $\alpha$, $\beta$ and $\gamma$. To avoid arbitrary optimization solutions, we constrain that $\alpha+\beta+\gamma=10$. 
We use pre-trained GPT-2 and BART weights from Hugging Face~\footnote{\url{https://huggingface.co/}}. 
The models were trained on Nvidia RTX2080-Ti GPUs over the training set for two epochs. Unless noted with ``training''/``train'', the reported model performance are based on the validation set.

\section{Results}
\label{sec:results}

\subsection{Overall Performance}
\label{sec:results:overall}

\begin{table*}[t!]

\centering
\caption{Comparison Between the Best Performing \mbox{\ourmethod} and Baseline \pkfocus}
\label{table:performance:compare}
\begin{threeparttable}
\bgroup
\def\arraystretch{1}%
\begin{tabular}{
  @{\hspace{0pt}}l@{\hspace{5pt}}|
  @{\hspace{5pt}}r@{\hspace{5pt}}
  @{\hspace{0pt}}r@{\hspace{5pt}}
  @{\hspace{0pt}}r@{\hspace{5pt}}
  @{\hspace{0pt}}r@{\hspace{5pt}}
  @{\hspace{0pt}}r@{\hspace{5pt}}
  @{\hspace{0pt}}r@{\hspace{5pt}}
  @{\hspace{0pt}}r@{\hspace{5pt}}
  @{\hspace{0pt}}r@{\hspace{5pt}}
  @{\hspace{0pt}}r@{\hspace{5pt}}
  @{\hspace{0pt}}r@{\hspace{5pt}}
  @{\hspace{0pt}}r@{\hspace{5pt}}
}
\toprule
                    Model   & F1   & ROUGE1 & ROUGE2 & ROUGEL & BLEU & PG(\%)  & PG\_MTL(\%) & KG(\%)  & PPL   & Inf. Time & Train. Time    \\  \midrule
BART + \pkfocus        & 0.291 & 0.353  & 0.186   & 0.311   & 11.364   & \bf{86.70} & \bf{37.21}    & 68.61 & 25.23 & \bf{2581.56} & 43645 \\
BART + \ourmethod (default) & 0.288 & 0.349   & 0.184   & 0.306   & 11.060   & \bf{86.70} & 37.19    & 69.87 & 14.61 & 2746.87 & \bf{26444} \\
BART + \ourmethod (search)$^{*}$     & \bf{0.317} & \bf{0.382}   & \bf{0.213}   & \bf{0.337}   & \bf{12.882}   & 86.69 & \bf{37.21}    & \bf{89.61} & \bf{13.17} & 2766.73 & 33110 \\ 
improvement (\%)    & 8.93  & 8.22    & 14.52   & 8.36   & 13.36    & 0.00 &  0.00     & 30.61 & 47.80 & -7.17 & 24.14 \\
\midrule
GPT2 + \pkfocus          & 0.261 & 0.338   & 0.173   & 0.302   & 9.923    & \bf{86.70} & \bf{37.19}    & 63.38 & 16.01 & \bf{1346.58} & 41340 \\
GPT2 + \ourmethod (default)$^{*}$ & \bf{0.279} & \bf{0.352}   & \bf{0.186}   & \bf{0.314}   & \bf{10.962}   & \bf{86.70} & \bf{37.19}    & 65.74 & \bf{10.97} & 1428.66 & 39406 \\
GPT2 + \ourmethod (search)    & 0.276 & 0.347   & 0.182   & 0.309   & 10.581   & \bf{86.70} & \bf{37.19}    & \bf{65.88} & 11.22 & 1438.05 & \bf{34699} \\ 
improvement (\%)    & 6.90  & 4.14    & 7.51    & 3.97   & 10.47     & 0.00 & 0.00     & 3.72  & 31.48 & -6.10 & 4.68 \\
\bottomrule
\end{tabular}

\begin{tablenotes}[flushleft]
    \setlength\labelsep{0pt}
    \footnotesize
    \item Values in \textbf{bold} represent the best performance of the corresponding metric among all methods. Models denoted with ``(default)'' use the default hyper-parameters ($\alpha/\beta/\gamma=1/1/10$). Models denoted with ``(search)'' are the best performing models from hyper-parameter search in Table~\ref{table:performance:search}. ``Improvement'' measures the percentage improvement of the best \ourmethod model (denoted by ``$^{*}$'') over the baseline \pkfocus.
\end{tablenotes}
\egroup 
\end{threeparttable}
\vspace{-10pt}
\end{table*}

The comparison of the best performance of \ourmethod against the baseline method is summarized in Table~\ref{table:performance:compare}. 
When using both BART and GPT2 as the base language model $LM$, \ourmethod is able to significantly outperform the baseline \pkfocus in various evaluation metrics. In specific, with BART as $LM$, \ourmethod achieves PPL=13.17 (47.8\% improvement over the baseline method), BLEU=12.882 (13.36\%), ROUGE1=8.22 (8.22\%) and KG=0.8931 (30.61\%). A similar trend can also be observed when using GPT2 as $LM$. 
This indicates the superiority of our method \ourmethod over the baseline method. When leveraging word similarity and contextual interaction in the knowledge grounding stage, the model is able to better capture low-level signals from specific keywords in various input sources, identify their relevance to the conversation context, and further assist the language modeling. However, we didn't observe any improvement based on PG accuracy from persona grounding. Detailed discussions will be provided in Section~\ref{sec:ablation:pg}.

\subsection{RQ1: Can the \texttt{NCLI} effectively extract relevant information and identify relevant persona/knowledge entries relevant to the conversation context?}
\label{sec:results:rq1}

\begin{table*}[t!]

\centering
\caption{Performance of \ourmethod with Hyper-Parameter Search}
\label{table:performance:search}

\begin{threeparttable}
\bgroup
\def\arraystretch{1}%
\begin{tabular}{
      @{\hspace{0pt}}c@{\hspace{5pt}}
      @{\hspace{0pt}}c@{\hspace{5pt}}
      @{\hspace{0pt}}c@{\hspace{5pt}}
      @{\hspace{0pt}}c@{\hspace{5pt}}|
      @{\hspace{5pt}}r@{\hspace{5pt}}
      @{\hspace{0pt}}r@{\hspace{5pt}}
      @{\hspace{0pt}}r@{\hspace{5pt}}
      @{\hspace{0pt}}r@{\hspace{5pt}}
      @{\hspace{0pt}}r@{\hspace{5pt}}
      @{\hspace{0pt}}r@{\hspace{5pt}}
      @{\hspace{0pt}}r@{\hspace{5pt}}
      @{\hspace{0pt}}r@{\hspace{5pt}}
      @{\hspace{0pt}}r@{\hspace{5pt}}
      @{\hspace{0pt}}r@{\hspace{5pt}}
}
\toprule
$LM$ & $\alpha$ (KG) & $\beta$ (PG) & $\gamma$ (LM) & F1   & ROUGE1 & ROUGE2 & ROUGEL & BLEU & PG (\%)   & PG\_MTL (\%) & KG (\%)   & PPL   & TIME    \\ 
\midrule
BART  & 2 & 2 & 6 & 0.287 & 0.351   & 0.186   & 0.309   & 11.076   & \bf{86.70} & 37.19    & 71.41 & 13.79 & 2749.90 \\
BART  & 2 & 4 & 4 & 0.283 & 0.348   & 0.182   & 0.307   & 10.945   & \bf{86.70} & 37.19    & 71.61 & 13.78 & 2704.76 \\
BART  & 2 & 6 & 2 & 0.292 & 0.352   & 0.181   & 0.307   & 11.040   & \bf{86.70} & 37.19    & 71.38 & 15.14 & \bf{2701.61} \\
BART  & 4 & 2 & 4 & 0.307 & 0.373   & 0.209   & 0.331   & 12.404   & \bf{86.70} & 37.19    & 87.07 & \bf{11.62} & 2781.97 \\
BART  & 4 & 4 & 2 & 0.282 & 0.344   & 0.181   & 0.303   & 10.863   & \bf{86.70} & 37.19    & 70.77 & 16.31 & 2782.25 \\
BART  & 6 & 2 & 2 & \bf{0.317} & \bf{0.381}   & \bf{0.213}   & \bf{0.337}   & \bf{12.882}   & 86.69 & \bf{37.21}    & \bf{89.61} & 13.17 & 2766.73 \\
\midrule
GPT2  & 2 & 2 & 6 & \bf{0.276} & \bf{0.347}   & \bf{0.182}   & \bf{0.309}   & \bf{10.581}   & \bf{86.70} & 37.19    & \bf{65.88} & \bf{11.22} & 1438.05 \\ 
GPT2  & 2 & 4 & 4 & 0.271 & 0.340   & 0.177   & 0.303   & 10.301   & \bf{86.70} & 37.19    & 65.01 & 11.32 & 1459.21 \\
GPT2  & 2 & 6 & 2 & 0.228 & 0.292   & 0.142   & 0.260   & 8.287    & \bf{86.70} & \bf{37.21}    & 56.82 & 11.96 & 1350.20 \\
GPT2  & 4 & 2 & 4 & 0.258 & 0.325   & 0.165   & 0.290   & 9.425    & \bf{86.70} & 37.19    & 61.16 & 12.28 & 1453.33 \\
GPT2  & 4 & 4 & 2 & 0.252 & 0.315   & 0.157   & 0.280   & 9.196    & \bf{86.70} & 37.19    & 62.56 & 12.87 & 1408.25 \\
GPT2  & 6 & 2 & 2 & 0.218 & 0.276   & 0.135   & 0.244   & 7.704    & \bf{86.70} & 37.19    & 59.86 & 13.84 & \bf{1342.84} \\
\bottomrule
\end{tabular}

\begin{tablenotes}[flushleft]
    \setlength\labelsep{0pt}
    \footnotesize
    \item Values in \textbf{bold} represent the best performance of the corresponding metric among all methods. To avoid trivial optimization solutions, we set a constraint that $\alpha+\beta+\gamma=10$.
\end{tablenotes}
\egroup
\end{threeparttable}
\vspace{-15pt}
\end{table*}



As reported in Table~\ref{table:performance:search}, we observe that BART generally outperforms GPT-2 in many performance metrics. Moreover, we find that (a) when using BART, 
assigning more weights to knowledge grounding (greater $\alpha$) will lead to better results, yet (b) when using GPT-2,  assigning more weights to the language model itself (greater $\gamma$) usually provides better results.
Therefore, our answer to {\bf RQ1} is: {\bf Knowledge grounding based on \texttt{NCLI} can effectively extract relevant information and identify relevant knowledge candidates. We also find that different language models require different hyper-parameter configurations and that fine-tuning the models can usually provide significantly better results.}

\subsection{{RQ2:} Performance-wise, is it helpful to adjust weights on different loss functions and use hyper-parameter optimization?}
\label{sec:results:rq2}

Table~\ref{table:performance:compare} compares the baseline models with default hyper-parameter settings against the best models selected from Table~\ref{table:performance:search}. For both BART and GPT2, we observe that the \ourmethod can obtain superior or on-par performance compared to the ones trained with default hyper-parameters.
Our method \ourmethod can significantly improve most metrics in both BART (e.g., 30.61\% in KG, 47.80\% in PPL) and GPT-2 (e.g., 3.72\% in KG, 31.48\% in PPL) scenarios.
Even though the best \ourmethod performance is achieved when using the default hyper-parameter, our ablation study (discussed later in Section~\ref{sec:ablation}) shows that there are clear trends that hyper-parameter $\alpha$ and $\gamma$ will significantly impact the model's performance in all evaluation metrics. 
Thus, our answer to {\bf RQ2} is: {\bf hyper-parameter optimization can generate much better models, and careful hyper-parameter search and choice are critical to the model performance.}

\subsection{{RQ3:} Is our method more computationally efficient?}
\label{sec:results:rq3}
As shown in Table \ref{table:performance:compare}, for training, the best performing \ourmethod models are able to improve total training time by 24.14\% and 4.68\% when using BART and GPT2 language models, respectively. This is mainly due to the fact that \ourmethod is able to reuse the language embeddings of persona/knowledge/utterance ($T[P]$/$T[K]$/$T[U]$) during the grounding processes, while the baseline method \pkfocus makes multiple $LM$ calls for each of the inputs. 
For testing inferences, the best performing \ourmethod models take about 6~7\% longer time than the baseline models. We point out that the testing inference time is mainly determined by the length of generated text when using the auto-regressive model. While our models are slightly slower than the baseline models, the improvement on the quality of the generated languages should take precedence over the longer reference time. 

More importantly, the framework provided by \ourmethod allows the persona/knowledge entry embeddings ($T[P]$/$T[K]$) to be pre-computed and cached, which could significantly improve the references time, while the baseline \pkfocus does not allow such caching because the utterances are unknown ahead of time. This flexibility to pre-compute and cache embeddings are critical to many real world applications where real-time model responses are expected on the scale of seconds or even milliseconds. 


In summary, our answer to {\bf RQ3} is: {\bf Our method is more efficient than the baseline methods at training time, and has the flexibility to significantly improve inference time through caching.}



\section{Ablation Study}
\label{sec:ablation}

\subsection{Language Model Choice: $LM$}
\label{sec:ablation:language-model}
    From Table~\ref{table:performance:compare} and~\ref{table:performance:search}, we observe that, when using BART as the base language model $LM$, the models are able to in general outperform the models that use GPT2 as the baseline $LM$. 
    Another interesting observation from our hyper-parameter search in Table~\ref{table:performance:search} is that models trained with BART and models trained with GPT2 have different trends w.r.t different hyper-parameters. For example, as the weight $\alpha$ on knowledge grounding increases, BART model performance gets better, but GPT2 model performance gets worse (more of these trends will be discussed in Section~\ref{sec:ablation:kg} to~\ref{sec:ablation:lm}).
    Our hypothesis is that this is probably due to the complexity of these two types of models. 
    BART as a recurrent model has 406M parameters, and GPT2 as a transformer model has 1.5B parameters. In addition, GPT2 models require all-to-all attention over all tokens in a text. These complexities made GPT2 extremely expensive to train - probably way beyond the two epochs in our experiment. Therefore, GPT2 performance naturally follows the weight on language model $\gamma$, and when increasing the PG weight $\beta$, the model is naturally ``distracted'' and not fully trained, which will result in lower performance. BART, on the other hand, is a simpler model and the language generation task can be relatively easily optimized and thus benefit more from better knowledge grounding. 
    As the main purpose of this paper is to show the effectiveness of our method \ourmethod with \texttt{NCLI}, we will leave the validation of such hypothesis in our future research. 

\subsection{Knowledge Grounding Weight: $\alpha$}
\label{sec:ablation:kg}
    \begin{figure}[t!]
        \centering
        \includegraphics[width=0.6\linewidth]{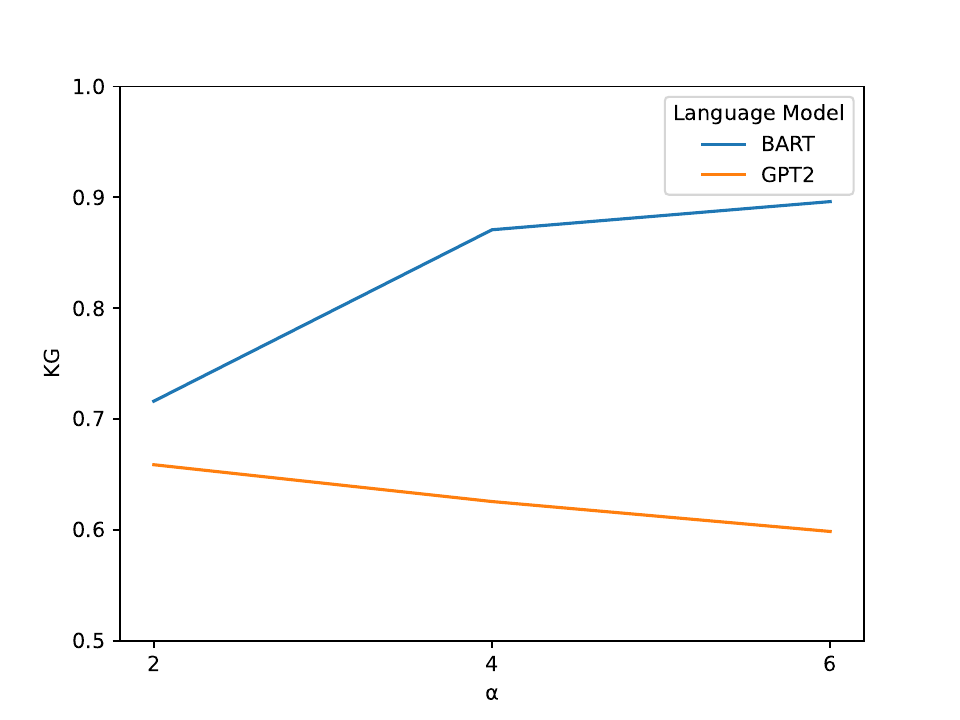}
        \caption{Effectiveness of $\alpha$ on KG}
        \label{fig:alpha-kg}
        \vspace{-10pt}
    \end{figure}    
    
    Figure~\ref{fig:alpha-kg} shows the best KG performance of \ourmethod w.r.t different $\alpha$ values. 
    We observe that, for BART, the KG performance is positively related to the choice of $\alpha$. When the weight increases on the knowledge grounding, the KG accuracy increases. This shows that the model is able to effectively extract signals from the knowledge candidates and the other context information and correctly identify a relevant knowledge entry from the candidates. 
    For GPT2, on the contrary, as discussed in Section~\ref{sec:ablation:language-model}, suffer from its complexity and were not able to learn sufficiently during the two epochs, therefore, increasing the KG weight would adversely affect the language modeling performance, and then affect the knowledge grounding, as knowledge grounding also takes $LM$ outputs as its inputs. 
    
    
\subsection{Persona Grounding Weight: $\beta$}
\label{sec:ablation:pg}
    From both Table~\ref{table:performance:compare} and~\ref{table:performance:search}, both PG and PG\_MTL accuracies remain unchanged in our experiments, regardless of changes in any hyper-parameters, including $\beta$. There are three potential underlying reasons. 1. The PG task is formulated as a multi-labeling classification task that identifies zero or more persona labels, which is naturally more difficult than regular single-label classification tasks like knowledge grounding and might require longer time and sufficient data to be trained well. 2. The PG labels are highly sparse and unbalanced in the dataset. False labels dominate the dataset, which make the learning problem even harder. 
    In addition two these two issues, another observation on the dataset is that many conversations may not demonstrate a strong relevance to persona and show how persona will affect the conversations. An example could be having a question $U$=``Where is this place?'', and the ground truth answer $a$=``This is \texttt{[place\_name]}, a place you like to visit'' was labeled as using certain persona entry, but in fact ``\texttt{[place\_name]}'' might have be extracted from the conversation context already (e.g., knowledge or utterance history), and the labeled persona entries do not help much on the rest of information in the answer.

\subsection{Language Model Weight: $\gamma$}
\label{sec:ablation:lm}
    \begin{figure}[!h]
        \vspace{-15pt}
        \centering
        \begin{subfigure}{0.49\linewidth}
          \centering
          \includegraphics[width=\textwidth]{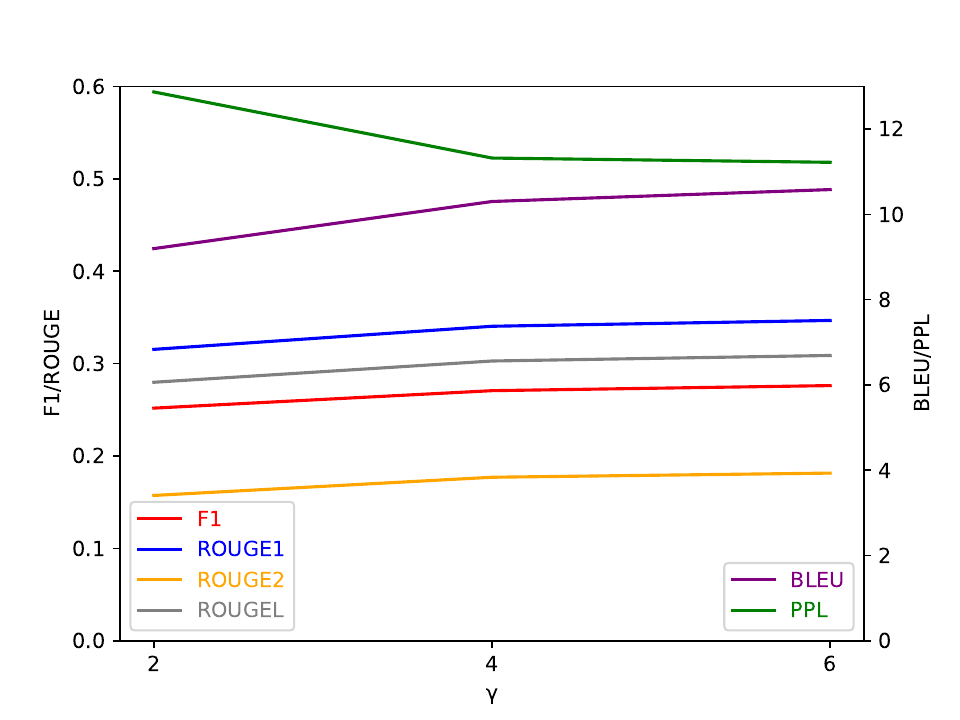}
          \vspace{-15pt}
          \caption{Effectiveness of $\gamma$ (GPT2)}
          \label{fig:gamma-gpt2}
        \end{subfigure}
        \begin{subfigure}{0.49\linewidth}
          \centering
          \includegraphics[width=\textwidth]{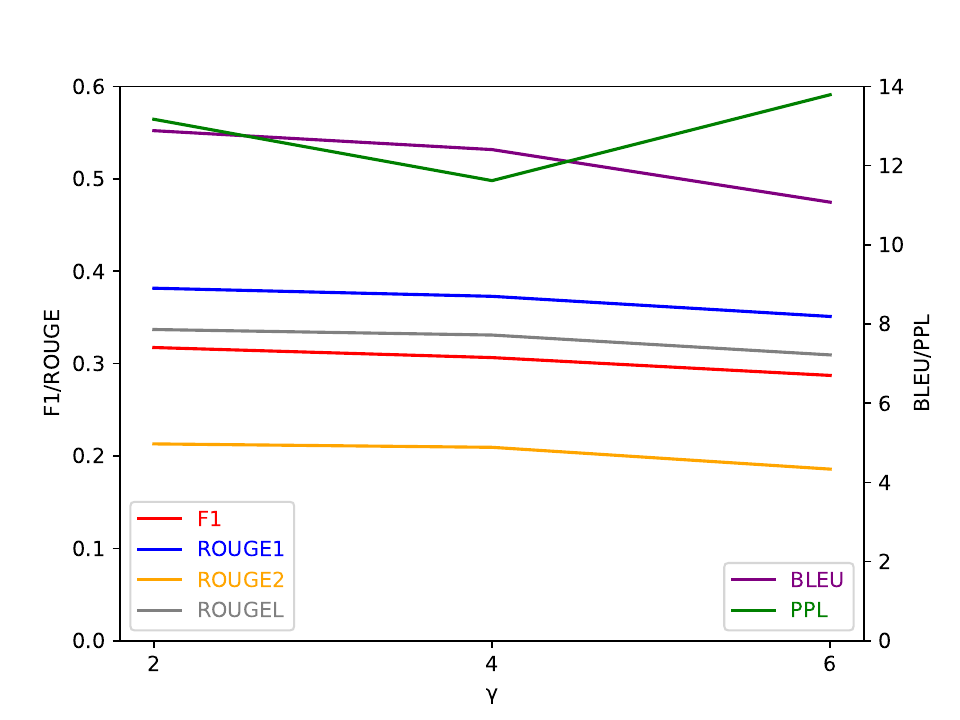}
          \vspace{-15pt}
          \caption{Effectiveness of $\gamma$ (BART)}
          \label{fig:gamma-bart}
        \end{subfigure}
        \caption{Effectiveness of $\gamma$ on $LM$}
        \label{fig:gamma}
    \end{figure}

    Figure~\ref{fig:gamma} shows the performance change on $LM$ w.r.t. different $\gamma$ values. 

    For GPT2, PPL drops as the value of $\gamma$ increases, which indicates the model benefits from a higher model weight setting for reducing the loss of language modeling, therefore, the model is more confident on the generated language and perplexity decreases. Meanwhile, the scores of Rouge1, Rouge2, RougeL, and BLEU have clear improvement with the higher $\gamma$ weights as well. 
    For BART, we also observe opposite trends that as $\gamma$ increases, the language evaluation performance decreases. BART is able to benefit more from identifying the correctly knowledge and leverage into the response generation stage. It finds all the potential candidates congregating all the user query inputs and pieces of knowledge. This is consistent with our hypothesis as discussed in Section~\ref{sec:ablation:language-model}.


\section{Conclusions}


Large language models and conversational agents based on AI are playing significant roles in many academic and commercial applications. It has become critical for the conversational agents to not only provide fluent responses but also respect knowledge and facts, and more importantly, personalize the responses to tailor to specific users' preferences. 
In this work, we presented our novel method, \ourmethod, a knowledge and persona grounding based model with normalized contextual latent interaction that is able to 1) identify persona and knowledge entries that are relevant to the conversation context, 2) generate high-quality responses, and 3) improve the computational efficiency over the state-of-the-art method \pkfocus. In specific, we designed a novel approach to persona and knowledge grounding via \texttt{NCLI}, and experimentally compared our method \ourmethod against \pkfocus. The experimental results suggested that \ourmethod has superior performance in terms of language quality and knowledge grounding, and achieved comparable performance in terms of knowledge grounding. We showed that \ourmethod has significantly better computational efficiency during training and is capable of being more efficient at inference time for deployed models via caching. We also studied how specific language model choice and different weights on persona grounding, knowledge grounding, and language modeling will affect the performance of \ourmethod. We also pointed out several directions for our future study, including exploring more effective ways to utilize user persona profiles for response personalization and studying how different language model attributes and specifications would affect the behavior of conversational agents. 


\bibliographystyle{IEEEtran}
\bibliography{main}

\end{document}